\documentclass[letterpaper, 10 pt, conference]{ieeeconf}  % Comment this line out if you need a4paper

\IEEEoverridecommandlockouts                              % This command is only needed if 
\usepackage{graphicx} % for pdf, bitmapped graphics files
\usepackage{subcaption}
\usepackage{amsmath} % assumes amsmath package installed
\usepackage{amssymb}  % assumes amsmath package installed
\usepackage{color}
\usepackage[normalem]{ulem}
\usepackage{comment}
\usepackage{flushend}
\def\RRR{\mathbb{R}}

\def\BB{\mathbf{B}}

\DeclareMathOperator{\Proj}{Proj}

\title{\LARGE \bf
Uncertainty-Aware Driver Trajectory Prediction at Urban Intersections }

\author{Xin Huang$^{1,2}$, Stephen G. McGill$^{1}$, Brian C. Williams$^{2}$, Luke Fletcher$^{1}$, Guy Rosman$^{1}$% <-this % stops a space 
\thanks{This work was part of X. Huang's internship at Toyota Research Institute (TRI). However, this article solely reflects the opinions and conclusions of its authors and not TRI or any other Toyota entity.}% <-this % stops a space
\thanks{$^{1}$Toyota Research Institute, Cambridge, MA 02139, USA {\tt \string{stephen.mcgill,luke.fletcher,guy.rosman\string}@tri.global}}%
\thanks{$^{2}$Computer Science and Artificial Intelligence Lab, Massachusetts Institute of Technology, Cambridge, MA 01239, USA
        {\tt \string{xhuang,williams\string}@csail.mit.edu }}%
\thanks{Video demonstration is available at https://youtu.be/clR08hRdtlM}
}

\begin{document}

\maketitle
\thispagestyle{empty}
\pagestyle{empty}

%%%%%%%%%%%%%%%%%%%%%%%%%%%%%%%%%%%%%%%%%%%%%%%%%%%%%%%%%%%%%%%%%%%%%%%%%%%%%%%%
\begin{abstract}

Predicting the motion of a driver's vehicle is crucial for advanced driving systems, enabling detection of potential risks towards shared control between the driver and automation systems. In this paper, we propose a variational neural network approach that predicts future driver trajectory distributions for the vehicle based on multiple sensors. 

Our predictor generates both a conditional variational distribution of future trajectories, 
as well as a confidence estimate for different time horizons. Our approach allows us to handle inherently uncertain situations, and reason about information gain from each input, as well as combine our model with additional predictors, creating a mixture of experts.

We show how to augment the variational predictor with a physics-based predictor, and based on their confidence estimations, improve overall system performance. 
The resulting combined model is aware of the uncertainty associated with its predictions, which can help the vehicle autonomy to make decisions with more confidence. 
The model is validated on real-world urban driving data collected in multiple locations. This validation demonstrates that our approach improves the prediction error of a physics-based model by 25\% while successfully identifying the uncertain cases with 82\% accuracy.

\end{abstract}

%%%%%%%%%%%%%%%%%%%%%%%%%%%%%%%%%%%%%%%%%%%%%%%%%%%%%%%%%%%%%%%%%%%%%%%%%%%%%%%%
\section{Introduction}

%% Themes
% urban X
% Temporal X
% Robust 
% Online
% statistics X

In the case of parallel autonomy systems, a human driver and an autonomous system share control of a vehicle. When the driver's actions will put the vehicle and its surroundings at risk, the autonomous system should intervene to avoid a calamity. In order to plan alternative actions for the car, the autonomous system requires knowledge of driver intention.
Trajectory prediction, then, is an important component in improving advanced driver-assistance systems (ADAS).

Several properties are required for successful and actionable prediction of a human driver's intent. Models must reason about the inherent uncertainty of the future trajectory in both the immediate and longer term. They must leverage all available sensory cues, and be able to reason about when the risk of the driver's control choices outweigh the risk of a system intervention.
%\CH{what does actionable mean?}\GR{actionable -- that is suitable for actions. We should drive this point throughout the paper all the way to the confidence part, since the confidence is mostly important if you have alternative of actions -- see BW's comment, it means we didn't get this point across.} 

Many existing trajectory prediction algorithms \cite{houenou2013vehicle, woo2017lane} output deterministic results efficiently. However, these methods fail to capture the uncertain nature of human actions. Probabilistic predictions are very useful in many safety-critical tasks such as collision checking and risk-aware motion planning. They can express both the intrinsically uncertain prediction task at hand (human nature) and reasoning about the limitations of the prediction method (knowing when an estimate could be wrong \cite{kendall2017uncertainties}).

\begin{figure}[t!]
    \centering
    \includegraphics[width=0.48\textwidth]{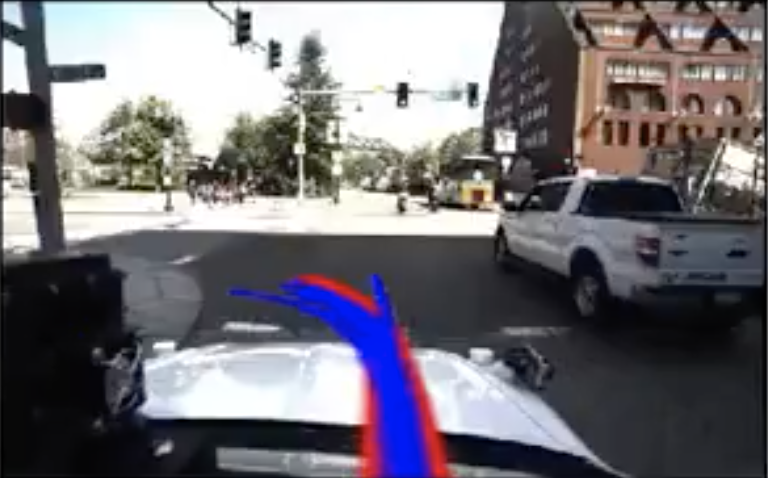}
    \caption{Illustration of a motivating example where a vehicle is in front of an intersection. The sampled predicted trajectories using our approach are plotted in blue, where the groundtruth future trajectory is plotted in red. In parallel autonomy, the autonomous system can leverage the predicted driver trajectory to avert risk and improve the driving. This requires the system to be confident of its predicted trajectories. }
   \label{fig:camera}
\end{figure}

To incorporate uncertainties into prediction results, data-driven approaches can learn common characteristics from datasets of demonstrated trajectories \cite{vasquez2009growing, wiest2012probabilistic, hermes2009long}. Unfortunately, these methods often express uni-modal predictions, which may not perform well in scenarios where the driver can choose among multiple actions. For instance, the vehicle stopping in front of the intersection in Figure~\ref{fig:camera} can move in two ways - turn left or go forward. To deal with such situations, our approach predicts multiple choices for the target vehicle.

Highway vehicle trajectory prediction \cite{park2018sequence, deo2018multi} represents a significant body of research, but work is lacking for urban driving prediction. Intersections, for instance, were responsible for 40\% of crashes happened in the United States in 2008 \cite{choi2010crash}. We therefore focus on predicting trajectories for vehicles driving in urban environments. This is more challenging than highway trajectory prediction due to more complicated environments with different road shapes and dynamic objects, as well as the variety of available driving actions for the driver. Additionally, in many cases it is crucial to be aware of the confidence of the prediction. In cases where those predictions cannot be accurately made, a later planning or parallel autonomy layer can take this into account, avoiding catastrophic outcomes due to mispredictions.

Kinematic and dynamic motion models are well studied, and they are efficient for motion predictions, especially over a short horizon \cite{houenou2013vehicle, berthelot2011handling}. However, due to unforeseen changes in control inputs, these models may fail over a longer horizon. Our approach arbitrates between a wheel odometry model that works well in simple predictions and a deep neural networks model that deals with more complicated cases. We do so by introducing a learned confidence estimator that predicts the effectiveness of each prediction method for different future timepoints, conditioned on the inputs.

The input features to our problem include on-board vehicle sensor data, while foregoing mapping information. While mapping information can be incorporated into driving predictions \cite{DBLP:journals/corr/abs-1811-10119}, mapping data can prove unreliable in many scenarios due to changing environments such as construction areas. We choose sensor inputs with an emphasis on invariance to changing environments so that our model can generate reliable predictions. Finally, using data from affordable and widely available sensors allows our algorithms to be deployed across a spectrum of vehicle platforms.
% demonstrate the robustness of our model on other publicly available dataset such as KITTI \cite{Geiger2013IJRR}.

Our work has three major contributions. First, we propose a variational trajectory prediction framework based on a deep neural network (DNN) with inputs from multiple sensor modalities, each of which contributes to different aspects of prediction, as we demonstrate.
Second, we combine this variational predictor with an alternative predictor, and choose the most confident outcome using a confidence measure. The confidence measure is learned through a separately trained DNN, and helps the final predictor achieve lower prediction error.
% , choosing the most confident outcome. We estimate the uncertainty of each predictor using a separately trained DNN, achieving large improvements in terms of prediction error.
Third, we quantify the prediction uncertainty with respect to the future time horizon. Knowing how far into the future a prediction can be trusted is important for choosing between different actions in shared-autonomy settings.

%Last, we show how our model achieves accurate results that outperform state-of-the-arts approaches in challenging urban conditions including intersections and turnarounds in multiple cities and under different weather conditions. \GR{Do we compare to other approaches? Or just our ablation study?}

The remainder of the paper is organized as follows: we give an overview of the existing work in Section~\ref{sec:related_works}, and formulate our problem in Section~\ref{sec:problem_formulation}. We then introduce the proposed method including a DNN-based variational trajectory predictor and a DNN-based confidence estimator in Section~\ref{sec:method}. Overall performance is presented in Section~\ref{sec:eval}, followed by a discussion on results and future work in Section~\ref{sec:conclusions}.

\subsection{Related Works}
\label{sec:related_works}

Several approaches exist for shared autonomy in terms of control and interfaces. The most common approach is to use advanced driver-assistance systems (ADAS) \cite{perez2015vehicle}, such as adaptive cruise control. Parallel autonomy vehicle shared-control frameworks, such as \cite{schwarting2017parallel, naser2017parallel}, produce safe trajectories, while minimizing the deviation from driver inputs. In order to minimize this deviation (as well as gauge risk), the system needs to know where the driver is going, which is a major focus of our work.

Many approaches have been used to predict vehicle trajectories \cite{lefevre2014survey}. Maneuver-based models \cite{houenou2013vehicle, deo2018multi, schreier2016integrated, tran2014online} employ a hierarchical approach, estimating high-level maneuvers and predicting vehicle trajectories for each maneuver type. With delineated maneuver classes, the predicted trajectories cover multiple actions of drivers. For example, Deo et al. \cite{deo2018multi} use a maneuver-based approach to predict multi-modal trajectories, and show promising results on highway driving data. Tran et al. \cite{tran2014online} learn maneuvers at intersections using Gaussian regression models, and, given the most likely class of maneuver, predict trajectories. Although these approaches afford multi-modal predictions, defining and labeling maneuvers can be challenging and time-consuming -- especially in urban driving with many road shapes and maneuver types.

% Different kinds of data have been used to predict future trajectories, including past trajectories \cite{houenou2013vehicle, kim2017probabilistic}, CANBUS data, images, IMU, and a combination of them. We decide to take advantage of all these data sources as they are very accessible and can be added to vehicles very easily to perform prediction tasks.

Some prediction approaches utilize temporal networks \cite{park2018sequence, deo2018multi, kim2017probabilistic, lee2017desire, alahi2016social} to capture the dynamics of the car and local environmental changes.
For example, long short-term memory (LSTM) networks \cite{park2018sequence, kim2017probabilistic} aid in predicting the grid-based vehicle locations probabilistically in horizons up to two seconds. 
Recently, \cite{lee2017desire} applies an RNN encoder-decoder framework with image and laser scan inputs to predict multi-modal trajectories in urban environments. The framework first proposes trajectory samples and then refines them with an inverse optimal control method.
In comparison, our approach focuses on producing predictions directly, without using a refinement scheme. 
For efficiency and robustness during training and testing, we capture the vehicle's past trajectory by a set of basis functions. As we show in the results section, using the past trajectory achieves comparable performance to RNN-based techniques.

By combining a short-term motion-based predictor and a long-term maneuver-based predictor, \cite{houenou2013vehicle} generates accurate trajectories over a horizon up to four seconds. Instead of setting a time-dependent weight function, our approach chooses weights based on estimated confidence over a range of prediction horizons. Our approach also allows us to gauge when the prediction result is not trustworthy, which is useful when considering whether to intervene, as in a shared-autonomy.

A crucial part of vehicle-related deep learning tasks is reasoning about the uncertainty of their outputs and knowing when the outputs are unreliable. A couple of works have addressed this concern using novelty detection on inputs.
For instance, \cite{richter2017safe} uses an auto-encoder based approach to detect novel inputs based on reconstruction error for a vehicle navigation task. \cite{amini2018variational} estimates pixel-wise uncertainties from the input vehicle image using the VAE encoder network. Additionally, \cite{ramanagopal2018failing} presents a system to detect temporal and stereo inconsistencies between similar image pairs in order to recognize vehicle perception failures. \cite{amini2018localization} use a variational network to reason about mapping and localization failures via marginalization. In our work, we focus on producing an uncertainty-aware predictor that recognizes cases where it is not confident in its output.

% Gaussian mixture model (GMM) has been used to represent multi-modal probabilistic predictions \cite{wiest2012probabilistic, deo2018would, havlak2014discrete}. These work models the positions at each time step as a GMM, where the number of unknown parameters grows linearly with prediction horizons. Our representation only consists of coefficients resulting from projecting trajectories onto polynomial basis, which is independent of the prediction horizons or time discretization. \CH{Do we need this?}

\section{Problem Formulation}
\label{sec:problem_formulation}

We consider the probability distribution of the future positions of a target vehicle, conditioned on its past trajectory and current sensor inputs. These inputs include camera images, CAN bus inputs from the steering and acceleration system, and velocities.

%In this section, we formulate the probabilistic motion prediction problem as an estimation of the probability distribution of the future positions of a target vehicle, conditioned on its past trajectory and a few sensor inputs at the prediction time including camera images, CANBUS inputs, and velocities.

\begin{figure}
    \centering
    \includegraphics[width=0.45\textwidth]{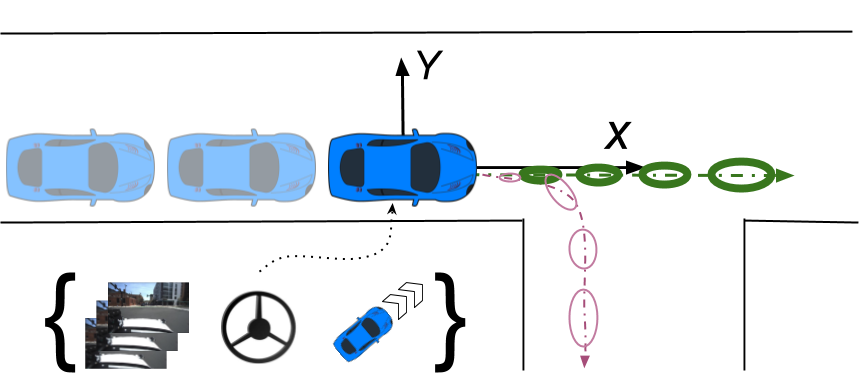}
    \caption{Illustration of the inputs and outputs in the proposed system. The inputs include past trajectories, camera images, CAN bus data, and IMU data. The outputs are the predicted vehicle positions as a multi-component Gaussian mixture model plotted in ellipses.}
    \label{fig:p_f}
\end{figure}

We leverage pose invariance by working in the local frame of the vehicle at the predicting time. As shown in Figure~\ref{fig:p_f}, the x-axis points to the heading of the vehicle, and the y-axis points to the left of the vehicle. All geometric quantities, such as positions and velocities, are transformed to the same frame. 

Similar to \cite{wiest2012probabilistic, deo2018would, havlak2014discrete}, we are assuming that the trajectories, projected onto some polynomial basis, form a Gaussian mixture model (GMM) with diagonal covariance matrices. Given a trajectory $\tau(t):[0,T] \rightarrow {\RRR^2}$ and function basis $\BB(t)$, the projection coefficients $c$ can be computed with Eq. \ref{eq:proj}. Reversely, given (sampling-dependent) basis and coefficients, the trajectory can be computed in Eq. \ref{eq:proj_rev}.
\begin{eqnarray}
    c =  \Proj_{\BB}(\tau) \label{eq:proj},\ \ \  \tau=\BB c \label{eq:proj_rev}.
\end{eqnarray}
This assume that the vehicle's localization is accurate enough, as was the case in our experiments.  
The basis projection operators are well defined regardless of the length of trajectory or sampling times, reducing the dimensionality of the input and output spaces for the learning problem.

The input $X$ to our model includes 
\begin{itemize}
\item Projected coefficients of past trajectory,
\item Steering wheel angle and gas pedal position,
\item Angular velocity and linear velocity,
\item Images from front camera and two side cameras.
\end{itemize}

The output of the model is a probability distribution over future trajectory, which can be transformed from a set of projected coefficients. Each coefficient is represented as a Gaussian mixture model. The number of components is predefined and used to represent the number of possible movements for the target vehicle. 

The goal is to estimate two conditional distributions $P(Y_{G}|X)$ and $P(Y_{C}|X)$. The first output $Y_{G}$ is the GMM parameters for the variational predictor including weights $w$, means $\mu$, and variances $\sigma$ of projected coefficients for the future trajectory. By modeling the prediction accuracy as a function of predicting horizon, the second output $Y_{C}$ contains a set of second-order polynomial coefficients that maps the time horizon to different confidence scores (e.g., L2 prediction error, root mean squared error) for each candidate trajectory predictor.

\section{Method}
\label{sec:method}

\begin{figure}
    \centering
    \includegraphics[width=0.48\textwidth]{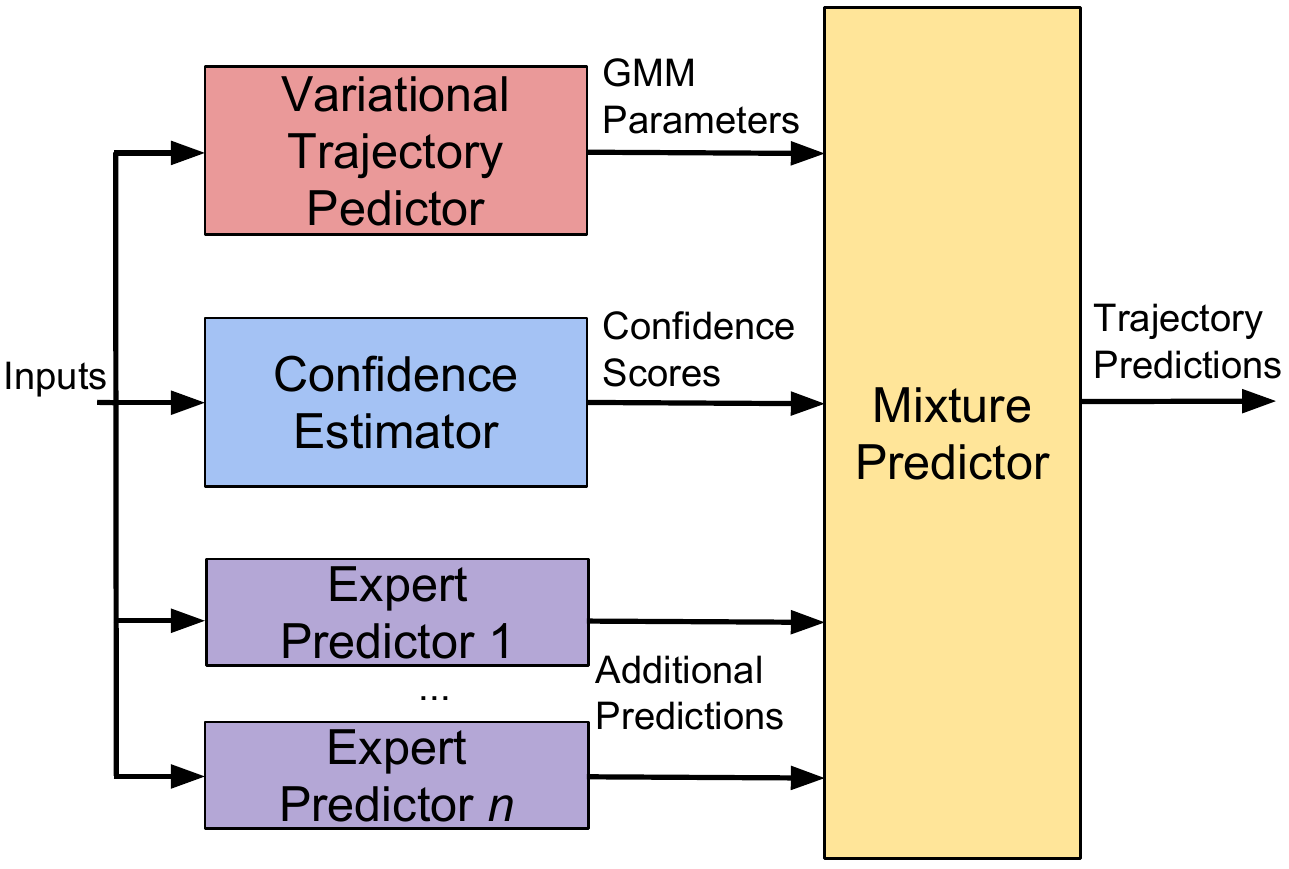}
    \caption{Architecture diagram of the proposed system, which includes a number of candidate trajectory predictors, a confidence estimator, and a mixture predictor generating final outputs.}
    \label{fig:arch_system}
\end{figure}

In this section, we describe our proposed predicting system depicted in Figure~\ref{fig:arch_system}. The system includes a variational trajectory predictor that outputs a set of GMM parameters, and a confidence estimator that generates confidence scores for the variational trajectory predictor as well as additional expert trajectory predictors. The mixture-of-experts predictor is responsible for choosing the best predictor among candidates according to the scores estimated by the confidence estimator, or providing a warning if none of the candidates are trustworthy. 

% \begin{figure}
%     \centering
%     \begin{subfigure}[b]{0.49\columnwidth}
%         \includegraphics[width=4.2cm]{figures/system_diagram.pdf}
%         \caption{}
%         \label{fig:arch_system}
%     \end{subfigure}
%     \begin{subfigure}[b]{0.49\columnwidth}
%       \includegraphics[width=4.2cm]{figures/trajectory_estimator.pdf}
%       \caption{}
%         \label{fig:arch_predictor}
%     \end{subfigure}
%     \caption{Architecture diagrams of the proposed system (a) and the variational trajectory predictor (b).
%     % \label{fig:architecture}
%     % , which includes a number of candidate trajectory predictors, a confidence estimator, and a mixture predictor generating final outputs.
%     }
% \end{figure}

\begin{figure}
    \centering
    \includegraphics[width=0.48\textwidth]{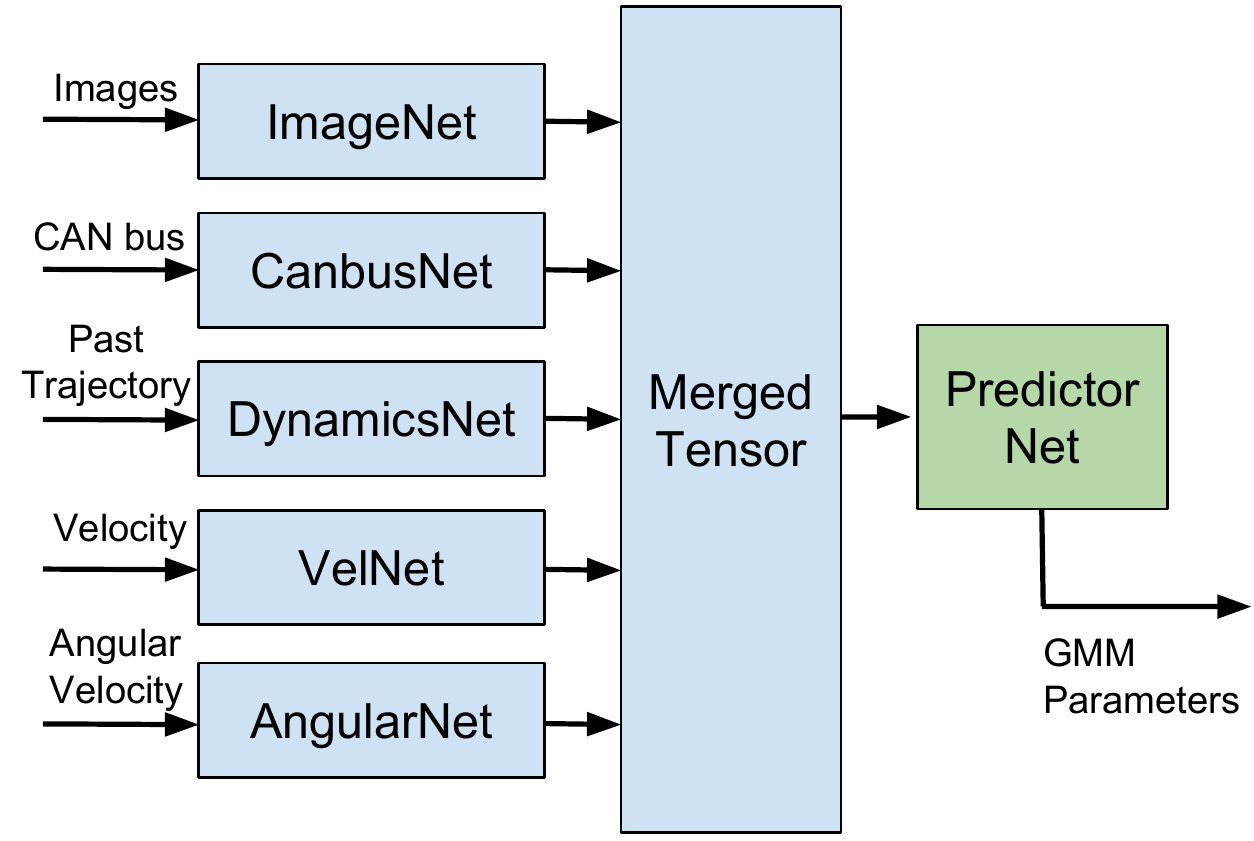}
    \caption{Architecture diagram of variational trajectory predictor, which is composed of a number of children networks processing different kinds of input data and a predictor network generating Gaussian mixture model parameters.}
    \label{fig:arch_predictor}
\end{figure}

\subsection{Variational Trajectory Predictor}
\label{subsec:trajectory_estimator}
The variational predictor model (see Figure~\ref{fig:arch_predictor}) is used to predict probabilistic coefficients as described in Section~\ref{sec:problem_formulation}. It consists of a number of children networks processing the sensor inputs and a predictor net that outputs the parameters of Gaussian mixture models.

\subsubsection{Children Networks}
\label{subsubsec:children}
Since the input data comes from various sensors and have different meaning and units, we implement a number of children networks specifically for each input, including a pre-trained VGG image network \cite{simonyan2014very} to process image data, and a set of networks for the rest children, consisting of a number of fully connected layers stacked on the top of each other with ReLU activations. Between fully connected layers we use a batch normalization layer to adjust and scale activation, and a dropout layer to prevent overfitting.

The outputs of each child network are concatenated together and sent to the predictor network. We add a block-dropout scheme at the concatenation that allows each child network to be ignored with a certain probability to avoid overfitting and to allow redundancy in the overall network.

\subsubsection{Predictor Network}
\label{subsubsec:predictor}

We use a variational prediction network to further process the information from all children. It consists of stacked fully connected layers with decreasing layer width, and dropouts layers in between. The network outputs GMM parameters, including weights, means, and (diagonal) covariances for each mixture component.

\subsection{Confidence Estimator}
\label{subsec:confidence_estimator}
The confidence estimator uses a similar structure to the variational trajectory predictor. It outputs a set of coefficients for different confident scores for each candidate predictor.

\subsection{Additional Expert Predictor}
\label{subsec:baseline_predictor}
We use an odometry-based predictor as the additional expert predictor that uses a wheel odometry model \cite{goldfain2018autorally} to compute the future positions of the vehicle by assuming the vehicle drives at constant turning rate and velocity.

Thanks to the modularity of our proposed framework, we can use more than one expert predictor. In this work, we only choose one for the sake of simplicity, and it performs well in complementing the variational predictor as discussed in Section~\ref{sec:eval}.

\subsection{Models training}
\label{subsec:train}
Each fully connected child network in Figure~\ref{fig:arch_predictor} has two hidden layers with a dimension of 10. The predictor network has four hidden layers where each of the first three layers has a dimension of 100 and the fourth layer has a dimension of 50. A batch size of 64 is used and the model is trained for 2000 epochs using Adam with a learning rate of 0.0001. Each child network has a dropout probability of 5\%. The confidence estimator network shares the same architecture and training procedures, but has a different output size and is trained with a different loss function. Both networks are implemented using PyTorch and trained on an AWS server with 4 Tesla V100 GPUs in parallel.

\subsubsection{Variational Trajectory Predictor Training}
The model is trained as in variational inference, with the loss function defined to be negative log-likelihood of the groundtruth trajectory coefficients given GMM parameters output by our model, in addition to several regularization terms.

The log probability for a single Gaussian component with basis dimensionality $D$ is computed using Eq. \ref{eq:lp}, and the negative log-likelihood for all mixture components is computed by Eq. \ref{eq:lq_all}.

\begin{eqnarray}
    LP_i(c) &=& \frac{1}{2}\sum_{d=1}^{D} -\log(2 \pi \sigma_d^2  ) - (c_d-\mu_d)^2/{\sigma_d^2} \label{eq:lp} \\
    \ell_{NLL}(c)&=&-\log{\sum_i w_i \exp{LP_i(c)}}, \label{eq:lq_all}
\end{eqnarray}
where $c_d$ are the individual projection coefficients of the trajectory.
To ensure the output weights and variance values are reasonable, we introduce an L2 loss on weight summation, an $L_{0.5}$ norm loss on individual weights, and an L2 loss on standard deviations. The total loss is a summation of individual losses described above.

% Additional regularization losses are introduced to ensure that the output weights and variances values are reasonable:
% \begin{itemize}
%     \item Weight-sum loss: $\ell_{W} = K_{W} * \big(\sum_i w_i - 1\big)^2$. This loss measures if the weights add up to 1.
%     \GR{either we explain we use weights in log space and normalize (this term prevents a loose DOF in the optimization), or remove this term as a detail.}
%     \item Component loss: $\ell_{C} = K_{C} * \|w\|_{1/2}= K_{C} * \sum_i w_i^{0.5}$. This loss add costs for low weights. We should say: $L_{0.5}$ norm.
%     \item Standard deviation loss: $\ell_{Std} = K_{Std} * \sum_i \sum_d (\sigma_{i,d} - \sigma_{desired}) ^2$. This loss measures the distance between the output variances and the desired values to prevent them from being too large or too small. \GR{Is it in log-space or not?}
% %     \item \sout{RNN hidden layers loss: $\ell_{RNN} = k_{RNN} * \sum_{k=1}^{L} |h_{k} - h_{k-1}|$, where $L$ is the number of sequence length in RNN. This loss measures the smoothness of hidden layer outputs from RNN.}l2_above_3_cnt/len(ours_end_l2s)
% \begin{comment}
%     \item Input reconstruction loss: $\ell_{recons} = \sum_i K_{i} ||P_i - O_i||^2$, which measures the L2 distance between predicted value and observed value for each input.
% \end{comment}

% \end{itemize}

\subsubsection{Confidence Estimator Training}
The loss function is defined as the L2 error between the predicted confidence scores computed using the coefficients output by our model and the actual confidence scores. The confidence score can be represented by any loss metric well-defined over the variational trajectory predictor and the expert trajectory predictor(s).

\section{Experimental Evaluation}
\label{sec:eval}

In this section, we start by describing the dataset used to train, validate, and test our model. We then show the contributions of each child network to the prediction results and describe the performance of the confidence estimator, followed by a discussion on the mixture predictor and its temporal performances.

\subsection{Data Collection and Processing}
\label{subsec:dataset}

\begin{figure}
\begin{subfigure}[b]{0.32\columnwidth}
    \includegraphics[width=2.8cm]{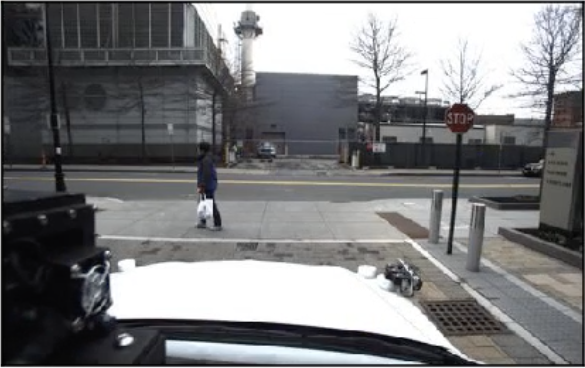}
    % \caption{Picture 1}
  \end{subfigure}
  \begin{subfigure}[b]{0.32\columnwidth}
    \includegraphics[width=2.81cm]{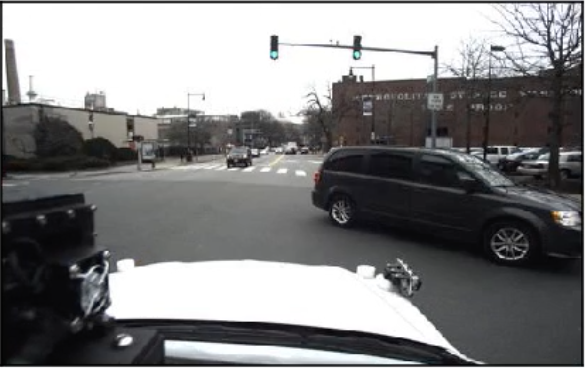}
    % \caption{Picture 2}
  \end{subfigure}
  \begin{subfigure}[b]{0.32\columnwidth}
    \includegraphics[width=2.82cm]{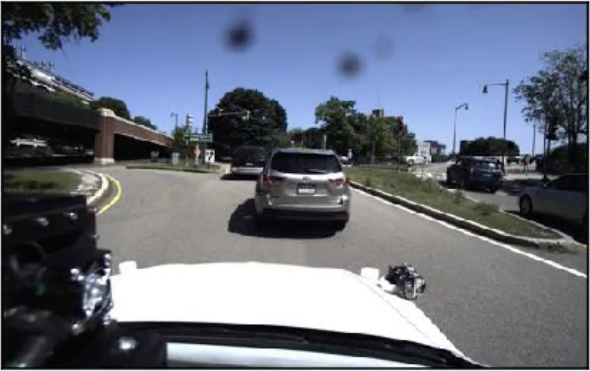}
    % \caption{Picture 2}
  \end{subfigure}
  \vspace{1mm}
\\
\begin{subfigure}[b]{0.32\columnwidth}
    \includegraphics[width=2.8cm]{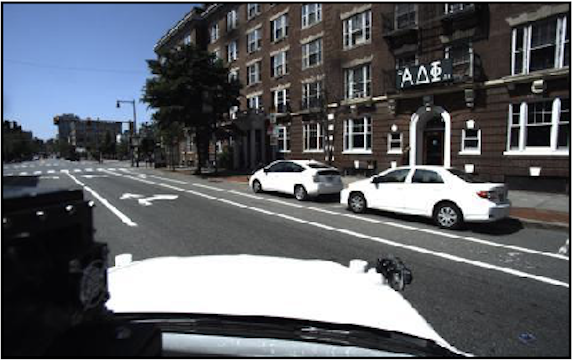}
    % \caption{Picture 1}
  \end{subfigure}
  \begin{subfigure}[b]{0.32\columnwidth}
    \includegraphics[width=2.81cm]{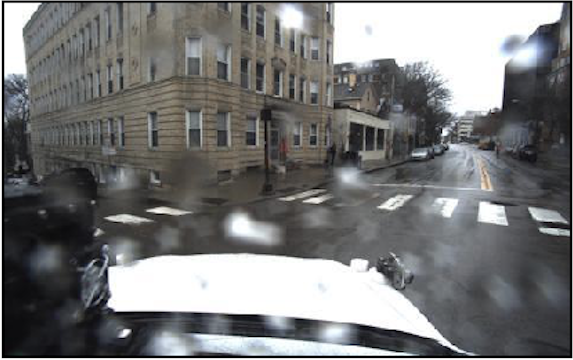}
    % \caption{Picture 2}
  \end{subfigure}
  \begin{subfigure}[b]{0.32\columnwidth}
    \includegraphics[width=2.82cm]{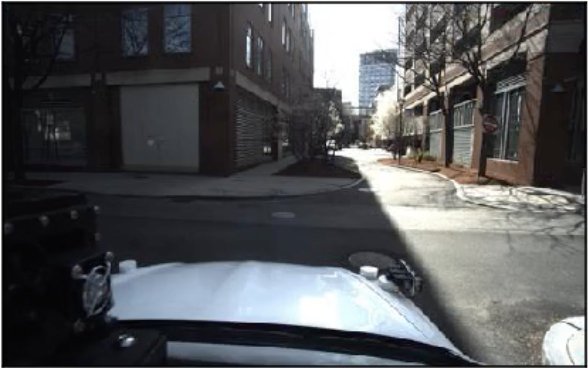}
    % \caption{Picture 2}
  \end{subfigure}
  \caption{Examples of image inputs encountered in our data. Top row: stop sign with a pedestrian, unprotected left turn, driving around roundabout. Bottom row: clear day weather, rainy weather, shady scenes.}
  \label{fig:env}
\end{figure}

Since we are mostly focused on intersection turning predictions, we decide to train and validate our model with driving data that are close to intersections (see Figure~\ref{fig:env}). The data was collected by a Toyota Lexus vehicle equipped with front and side cameras, Global Positions System (GPS), and Inertial Measurement Unit (IMU). It consists of 14 trips with a total duration of 30 hours in Cambridge MA, Ann Arbor MI, and Los Altos CA. Different weather and lighting conditions (see Figure~\ref{fig:env}) are included in the dataset, which is partitioned into a training and validation set and a testing set. The training and validation set is split arbitrarily twice for training the variational trajectory predictor and the confidence estimator, respectively. All the metrics mentioned in this section are computed on the same testing set.

\subsection{Children Networks}
\label{subsec:children}

In this part, we show the complementary contributions from children networks in terms of both statistics and specific examples. In order to quantify the performance statistically, we measure the negative log-likelihood (NLL) to estimate the information gain $I$ from the model. Information gain, or mutual information, for a specific input is given by the difference in conditional entropy with and without that input \cite{Cover:2006:EIT:1146355}. In our case this becomes the difference in log-probabilities over all data:

{\footnotesize
\begin{eqnarray}
I(\tau ; x)&=&H(\tau |{X^ - }) -H(\tau |{X^ - },x)\\
\nonumber
&=&-\hspace{-0.2cm}\sum\limits_{\substack{\tau,X^-,x}}  {p(\tau,X^-,x)\left[ {\log p(\tau |{X^-})} -\log p(\tau|{X^-} ,x) \right]}
\end{eqnarray}
}%
where $H(\cdot)$ is the entropy, $x$ denotes an input of interest, $X^{-}$ denotes the rest of the inputs, and $\log p(\tau|{X^ - },x),\quad \log p(\tau|{X^ - })$ are obtained by networks trained with or without $x$ as an input.

In addition, we use root mean squared error (RMSE) to show the quality of prediction along the entire future trajectory and L2 error at the predicting horizon of 3 seconds. To understand when the model fails catastrophically, we compute the percentage that the output predictions have an L2 error greater than 5 meters. The results are recorded in Table \ref{tbl:children_results}. 

\begin{table}[t!]
\centering
\begin{tabular}{|l|l|l|l|l|}
\hline
Model Name          & $I$ & RMSE [m] & L2 [m] & Hard \% \\ \hline \hline
Variational Predictor      &   N/A    &  1.15   &  3.01 & 4.83   \\ \hline
w/o Past Dynamics      &    0.81   &   1.29   & 3.37 & 8.43   \\ \hline
w/o CAN bus      &   0.73    &  1.28   & 3.41 & 10.24  \\ \hline
w/o IMU        &   0.85  & 1.31   &  3.45 & 10.47 \\ \hline
w/o Image       &   0.74    &  1.34   &  3.64 & 15.57 \\ \hline
\end{tabular}
\caption{Statistical performance of children networks. Left-to-right columns: information gain, root mean squared error, L2 endpoint error at 3 seconds, percent of hard cases (cf. text).}
\label{tbl:children_results}
\vspace{-4mm}
\end{table}

\subsubsection{Dynamics Network}
\label{subsubsec:dynamics}
\begin{figure}
\centering
\begin{subfigure}[b]{0.32\columnwidth}
    \includegraphics[width=2.79cm]{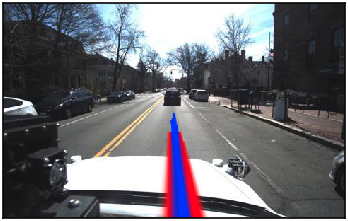}
  \end{subfigure}
  \begin{subfigure}[b]{0.32\columnwidth}
    \includegraphics[width=2.81cm]{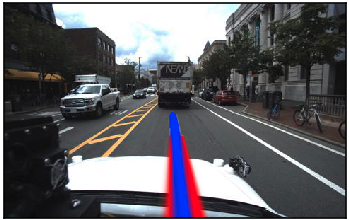}
  \end{subfigure}
\begin{subfigure}[b]{0.32\columnwidth}
    \includegraphics[width=2.77cm]{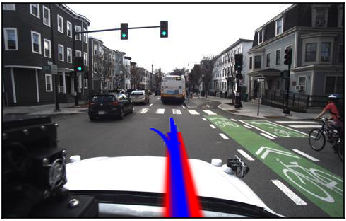}
  \end{subfigure}
(a) Without dynamics inputs, the predicted trajectories overshoot even when the vehicle drives on a straight road.\\
\vspace{1mm}
\begin{subfigure}[b]{0.32\columnwidth}
    \includegraphics[width=2.8cm]{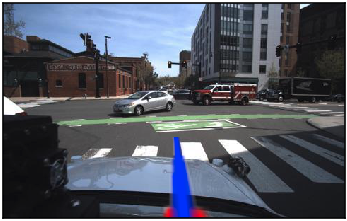}
  \end{subfigure}
  \begin{subfigure}[b]{0.32\columnwidth}
    \includegraphics[width=2.81cm]{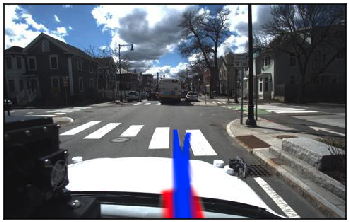}
  \end{subfigure}
\begin{subfigure}[b]{0.32\columnwidth}
    \includegraphics[width=2.8cm]{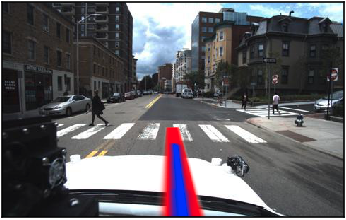}
  \end{subfigure}
(b) Without CAN bus inputs, there is always motion predicted even when the vehicle is not moving. \\
% \vspace{1mm}
% \begin{subfigure}[b]{0.32\columnwidth}
%     \includegraphics[width=2.8cm]{figures/no_imu_2.png}
%   \end{subfigure}
%   \begin{subfigure}[b]{0.32\columnwidth}
%     \includegraphics[width=2.8cm]{figures/no_imu_4.png}
%   \end{subfigure}
% \begin{subfigure}[b]{0.32\columnwidth}
%     \includegraphics[width=2.78cm]{figures/no_imu_5.png}
%   \end{subfigure}
% (c) Without IMU inputs, there are overshooting and undershooting predictions even when in smooth motion. \\
\vspace{1mm}
\begin{subfigure}[b]{0.32\columnwidth}
    \includegraphics[width=2.8cm]{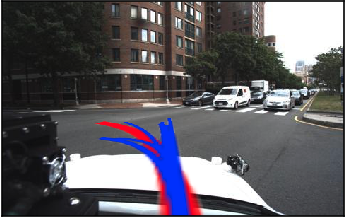}
  \end{subfigure}
  \begin{subfigure}[b]{0.32\columnwidth}
    \includegraphics[width=2.81cm]{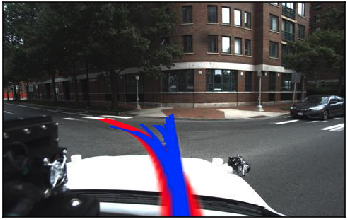}
  \end{subfigure}
\begin{subfigure}[b]{0.32\columnwidth}
    \includegraphics[width=2.8cm]{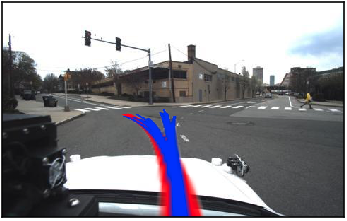}
  \end{subfigure}
(c) Without image inputs, the predicted trajectories go out of road boundaries especially at turns. \\
\vspace{1mm}
\begin{subfigure}[b]{0.32\columnwidth}
    \includegraphics[width=2.8cm]{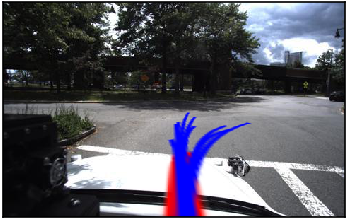}
  \end{subfigure}
  \begin{subfigure}[b]{0.32\columnwidth}
    \includegraphics[width=2.8cm]{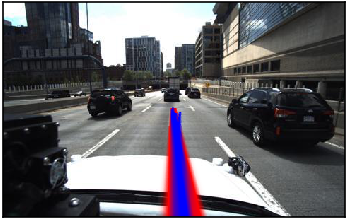}
  \end{subfigure}
\begin{subfigure}[b]{0.32\columnwidth}
    \includegraphics[width=2.79cm]{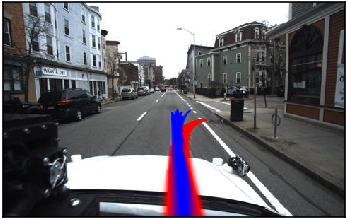}
  \end{subfigure}
(d) With all sensor inputs, the prediction can still be hard in less familiar or out-of-distribution scenarios.
\caption{Illustration of challenging examples for the variational predictor while withholding different inputs. The groundtruth future trajectory is in red, and the sampled predicted trajectories using our approach are in blue. }
  \label{fig:hard_examples}
\vspace{-2mm}
\end{figure}

We start by removing the dynamics network from the variational predictor model. Since we are using a second-order polynomial basis, the past coefficients capture the dynamics including the average past velocity and acceleration of the vehicle in each dimension. To check the contribution of the dynamics network, we visualize a few examples in Figure~\ref{fig:hard_examples}a where the prediction error increases drastically after removing the dynamics network from the variational predictor. We see that without knowing the past dynamics, the predictions can be slightly off when a vehicle drives smoothly on the road.

\subsubsection{RNN without Dynamics Network}
\label{subsubsec:dynamics}
As stated in Section~\ref{sec:related_works}, many use RNN-based approaches to learn temporal information such as dynamics from past observations. We implement an RNN model that takes the same input except for the information on the past trajectory. The model achieves an information gain of 0.17 as compared to the variational network without past dynamics, which is worse than the information gain of 0.81 we would gain from the past dynamics. This indicates that the dynamics network captures the majority of the signal that RNN-based models can learn. Given that the past trajectory is usually available for vehicles with GPS sensors, we decide to use a single shot model for our variational predictor, which takes less time to converge during training time than an RNN model.  We therefore do not show an RNN-based approach in the rest of the experiments and in Table~\ref{tbl:children_results}.

\subsubsection{CANBUS Network}
\label{subsubsec:canbus}
CAN bus data includes steering wheel angle and gas pedal value, which can inform the variational predictor about the instant acceleration and turning rate. From Table \ref{tbl:children_results}, we see that after dropping the CANBUS network, the L2 prediction error increases by 13.29\%. Without a CANBUS network, the variational predictor would not able to know the change in acceleration. Examples in Figure~\ref{fig:hard_examples}b show that at intersections, it is important to know the acceleration to avoid overshooting or undershooting in the predictions.

\subsubsection{IMU Network}
\label{subsubsec:imu}
In complementary to the CANBUS network, the IMU network provides information about the instant velocity and angular rate of the vehicle. 
% In Figure~\ref{fig:hard_examples}(c), we see that the IMU network is also crucial in preventing overshooting and undershooting in predicting trajectories.

\begin{figure}[t!]
\centering
\begin{subfigure}[b]{0.49\columnwidth}
    \includegraphics[width=4.2cm]{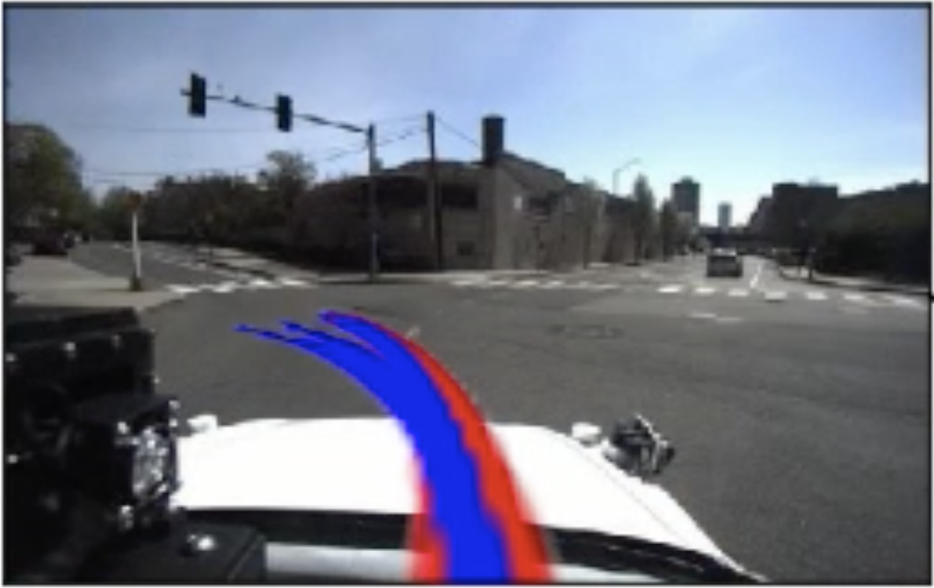}
  \end{subfigure}
\begin{subfigure}[b]{0.49\columnwidth}
    \includegraphics[width=4.2cm]{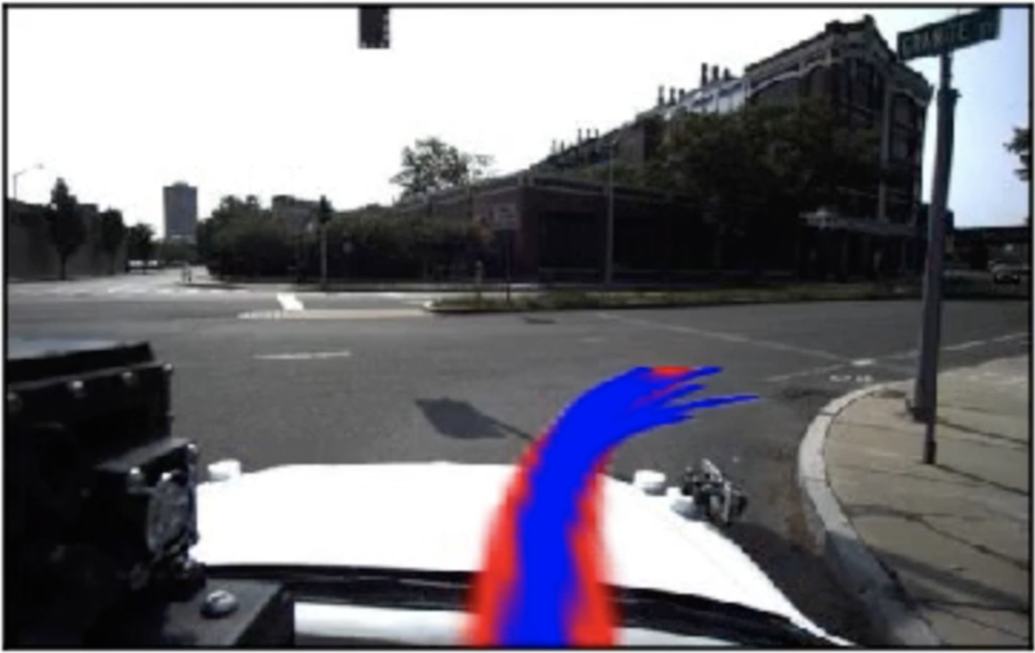}
  \end{subfigure}
\\
\vspace{1mm}
  \begin{subfigure}[b]{0.49\columnwidth}
    \includegraphics[width=4.2cm]{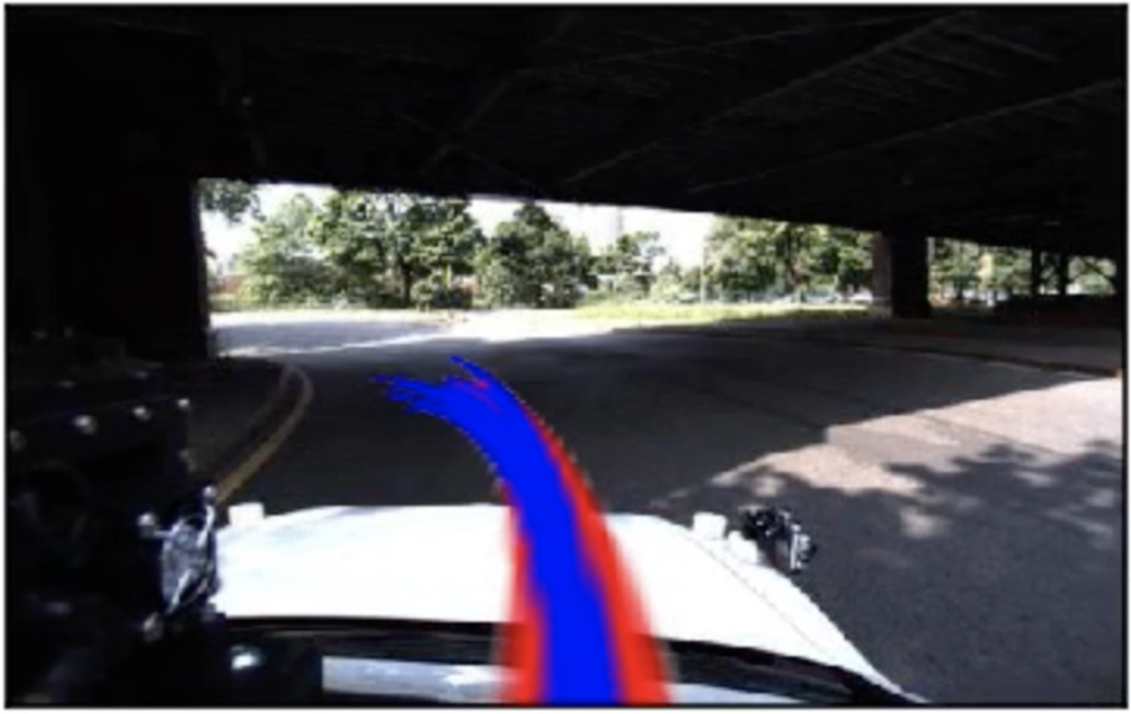}
  \end{subfigure}
\begin{subfigure}[b]{0.49\columnwidth}
    \includegraphics[width=4.2cm]{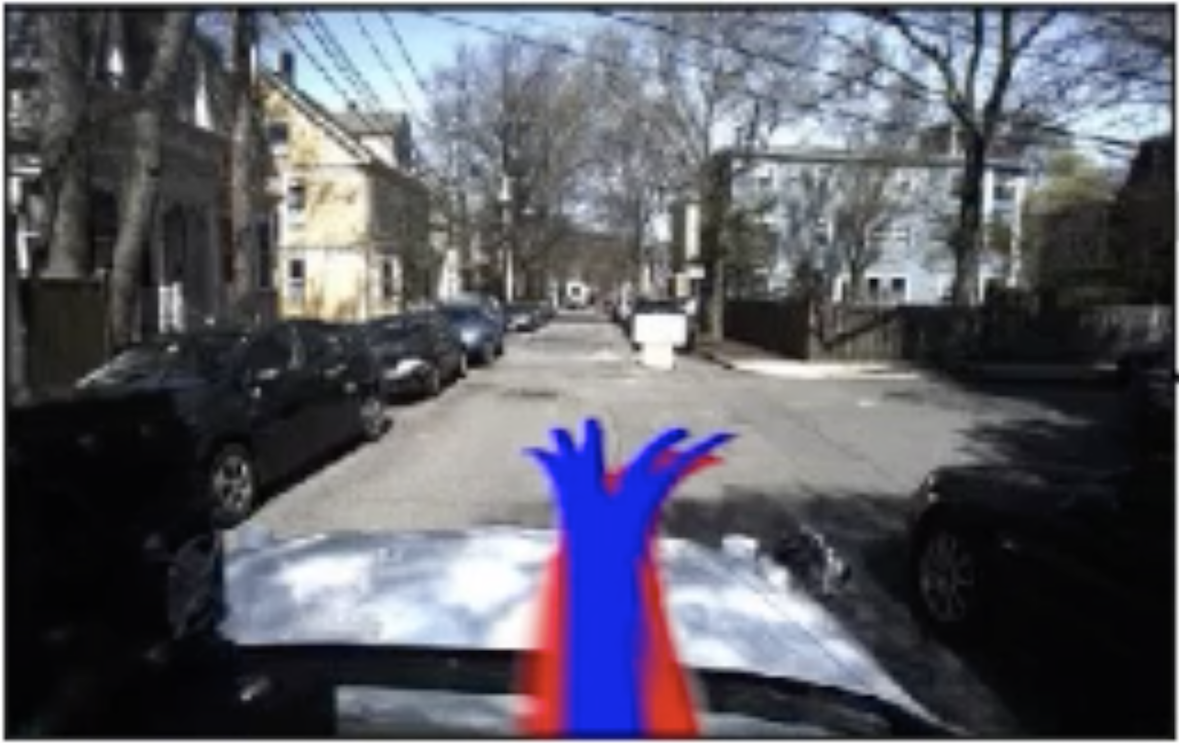}
  \end{subfigure}
\caption{Demonstrated predicted trajectories with the variational predictor. Top row: the vehicle is turning left and, and turning right. Bottom row: the vehicle is following roundabout beneath a highway, and facing two options in front of a T intersection. }
  \label{fig:successful_examples}
\vspace{-4mm}
\end{figure}

\subsubsection{Image Network}
\label{subsubsec:image}
The large performance drop after removing the image network suggests that it contributes the most to the network. For example, the percentage of hard cases has increased by two times if we remove the image network, which means that the image network is able to reduce many hard cases by using its information on what exists in the future, especially in cases where the past dynamics or the CAN bus/IMU data is unavailable. More specifically, the pre-trained VGG network allows us to identify the upcoming environment conditions such as road shapes and obstacles. This can help to fine-tune the predicted trajectories to be close to the road curve and yield higher accuracy. As shown in Figure~\ref{fig:hard_examples}c, without an image network, the variational predictor has a hard time getting accurate predictions when the vehicle turns at intersections.

\begin{figure*}[t!]
\begin{subfigure}[b]{0.48\linewidth}
\centering
\includegraphics[width=0.88\columnwidth]{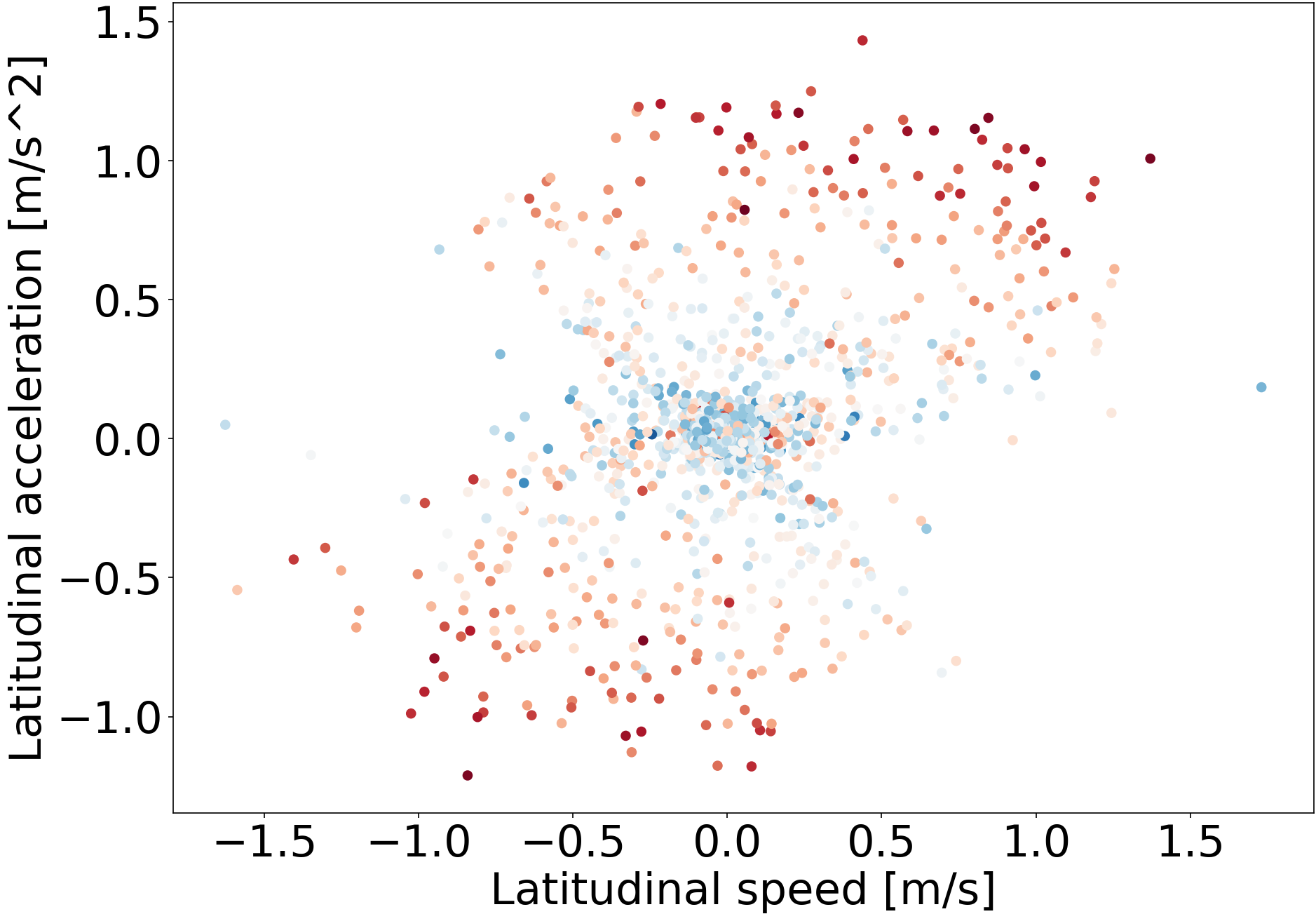}
\end{subfigure}
% \vspace{1mm}
% \\
\begin{subfigure}[b]{0.48\linewidth}
  \centering
    \includegraphics[width=0.88\columnwidth]{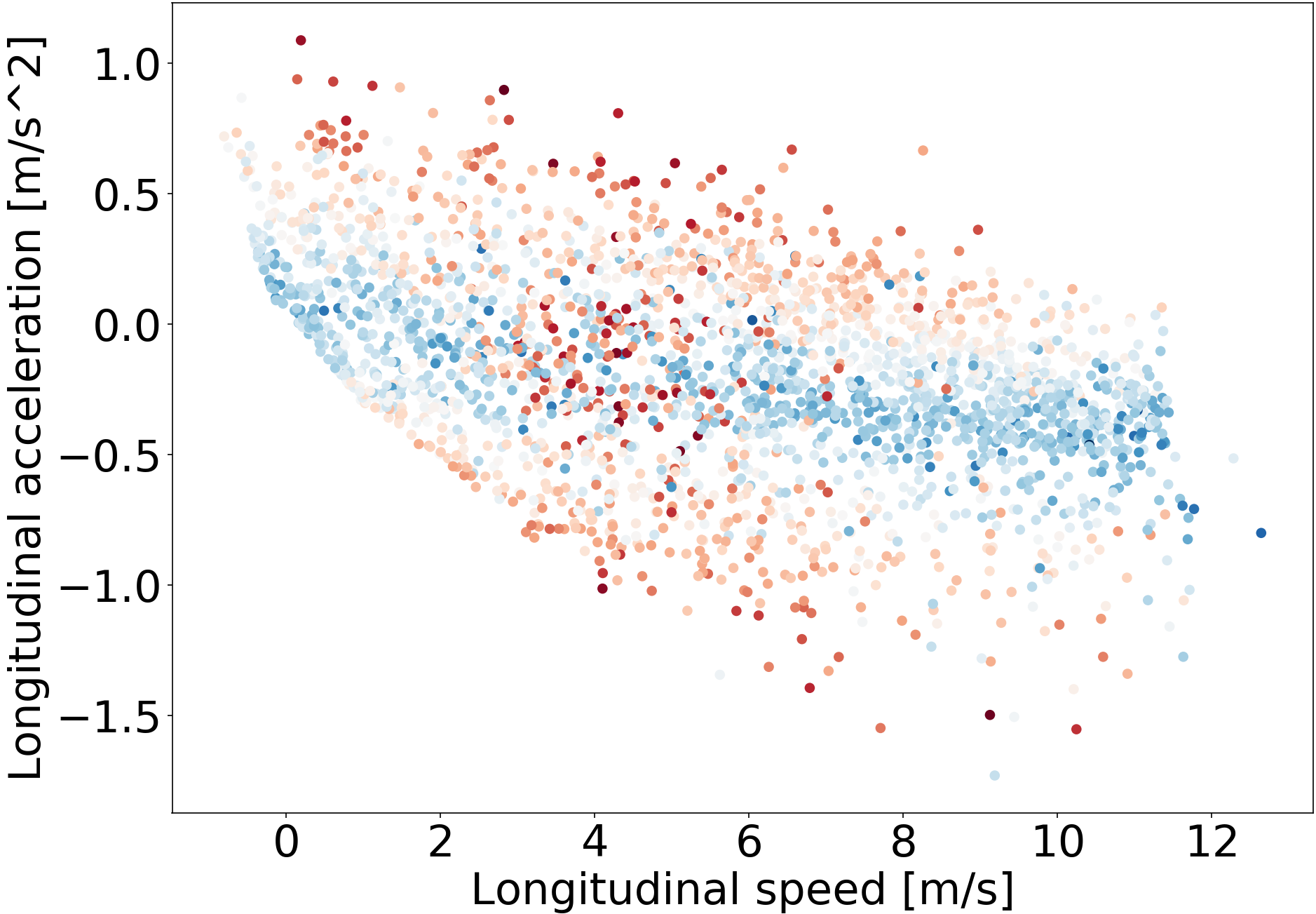}
  \end{subfigure}
\caption{Scatter plots of the L2 error differences between our variational trajectory predictor and the odometry predictor over 3 seconds. Red means the variational predictor performs better. Blue means the odometry predictor performs better. Top: Error differences as a function of latitudinal vehicle velocity and acceleration. Bottom: Error differences as a function of longitudinal vehicle velocity and acceleration. For qualitative results, please refer to our video demonstration.
% at https://youtu.be/clR08hRdtlM.
}
  \label{fig:l2_comparison}
\vspace{-2mm}
\end{figure*}

\begin{figure*}[t!]
\begin{subfigure}[b]{0.33\linewidth}
\centering
\includegraphics[height=1.7in]{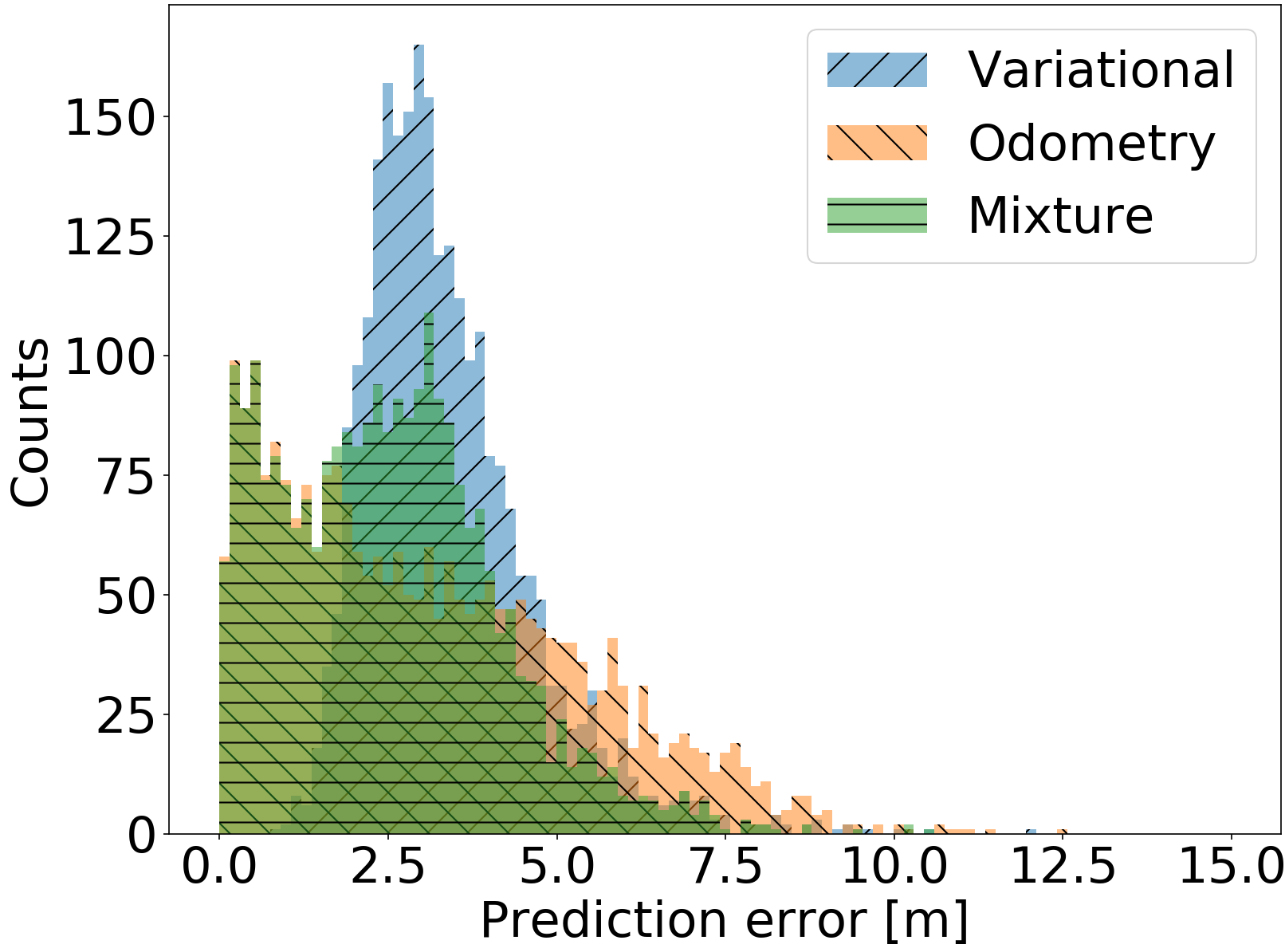}

(a)
\end{subfigure}
\begin{subfigure}[b]{0.33\linewidth}
  \centering
    \includegraphics[height=1.7in]{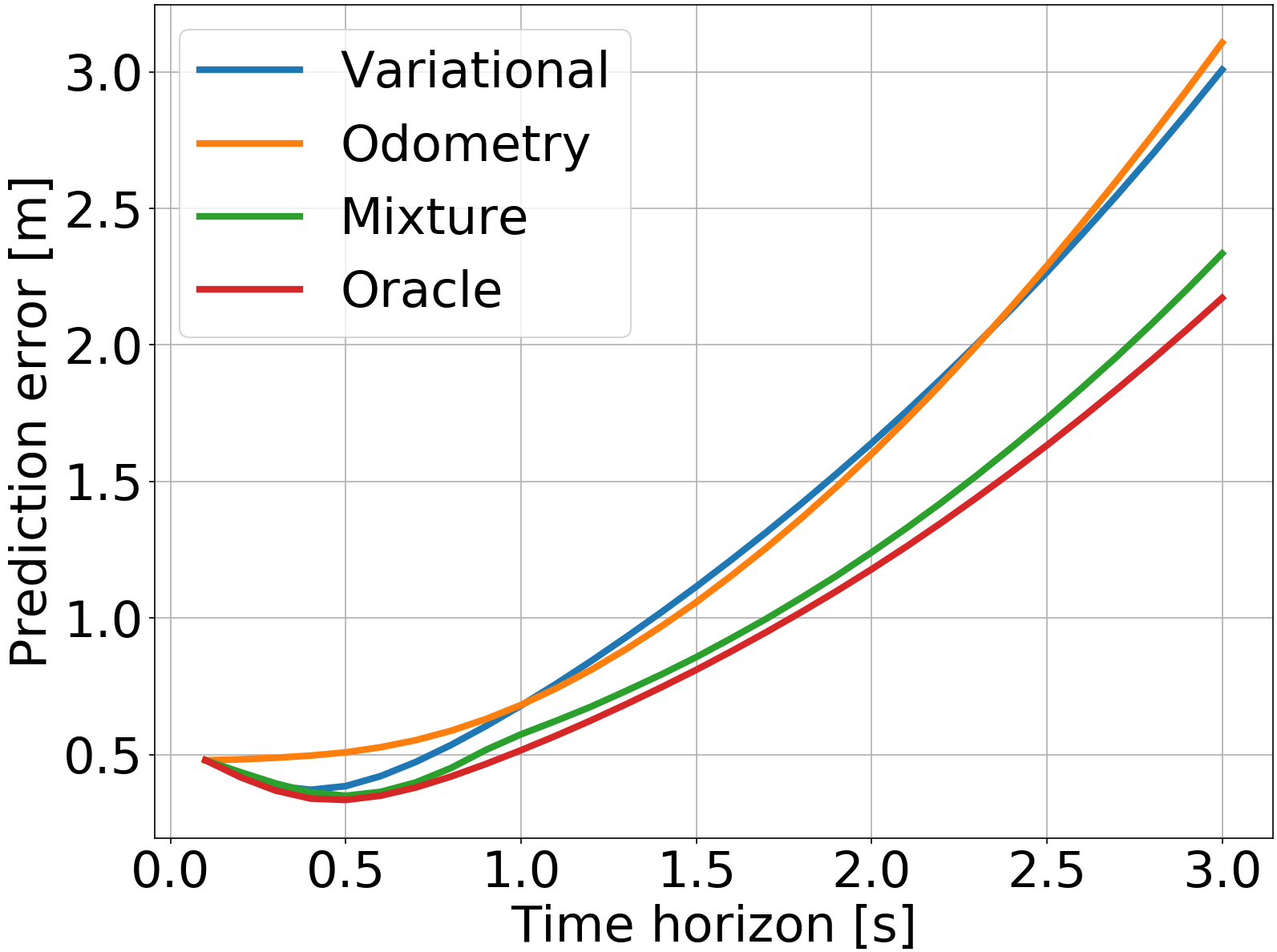}
    
    (b)
  \end{subfigure}
  \begin{subfigure}[b]{0.33\linewidth}
  \centering
    \includegraphics[height=1.7in]{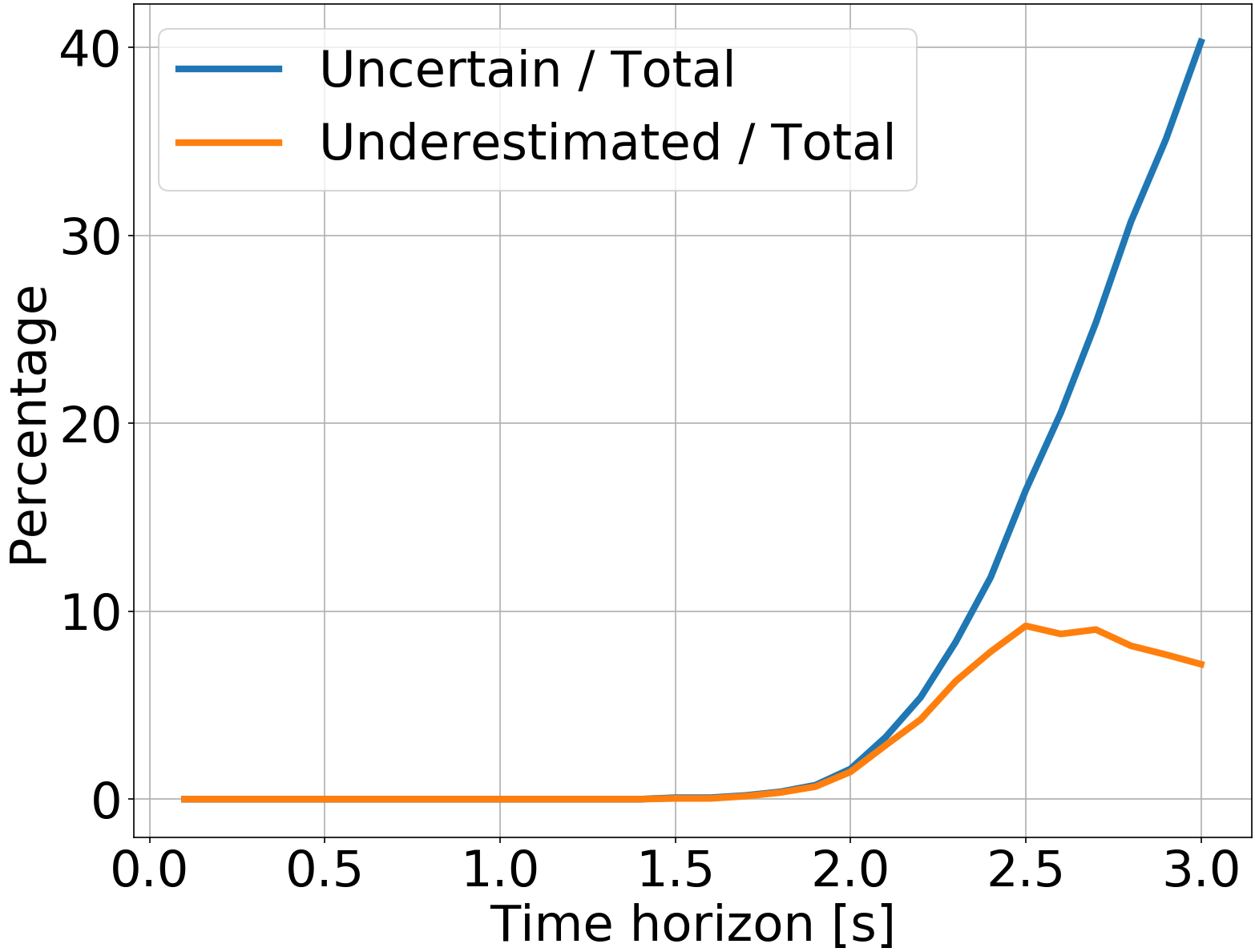}
    
    (c)
  \end{subfigure}
\caption{Illustrations of predictors performances on the testing set. (a) Histograms of prediction errors for the variational predictor, the odometry predictor, and the mixture predictor at 3 seconds.  (b) Prediction errors of three predictors and an oracle predictor, over different time horizons from 0.1 to 3 seconds. (c) Percentage of uncertain cases (cf. text) and underestimated uncertain cases over different time horizons.}
  \label{fig:temporal_trends}
\vspace{-4mm}
\end{figure*}

\subsubsection{Variational Trajectory Predictor}
\label{subsubsec:variational}
By using all the children networks, the variational predictor gets information about the environment ahead of the vehicle as well as the vehicle state including acceleration, velocity, angular rate, and past dynamics. It is able to predict trajectories with an average prediction error of 3.01 meters in complicated urban driving environments. Some successful predictions are illustrated in Figure~\ref{fig:successful_examples}.

Despite the success of the variational predictor, it can fail. Defining failures as estimation errors greater than 5 meters over the horizon, we have a 4.83\% failure rate in urban driving conditions. Figure~\ref{fig:hard_examples}d shows our failure examples in challenging scenarios including places without well-defined road boundary, in an unfamiliar highway environment, and in front of a street hidden behind the buildings. For a trajectory predictor to be trustworthy in a shared-autonomy framework, it has to recognize these uncertain cases, which motivates our confidence estimator, as described in the next section.

\subsection{Confidence Estimator}
\label{subsec:conf_est}
We use the results from the two candidate trajectory predictors to obtain confidence scores for each data sample. In this work, we choose to use the L2 error between the predicted and groundtruth position as our confidence score. The error distributions at predicting horizon of 3 seconds for two candidate predictors are shown in Figure~\ref{fig:temporal_trends}a, with an average of 3.01 meters and 3.11 meters, respectively. 

In order to compare the performance between the two candidate predictors at different scenes, we show in Figure~\ref{fig:l2_comparison} the scatter plots of L2 errors between the variational predictor and odometry predictor as a function of vehicle states, where red means the former performs better and blue means the latter performs better. We find that when the lateral accelerations are small, the odometry predictor performs better, as the vehicle is moving forward or turning smoothly without sudden lateral velocity changes. Similarly, in scenarios where longitudinal accelerations are small, which means the vehicle is driving at a constant velocity, the odometry predictor has lower errors than the variational predictor, as expected. On the other hand, when vehicle is driving at either large latitudinal or longitudinal accelerations, our variational predictor performs better in terms of L2 errors.

The obtained confidence scores are used to train a confidence estimator that produces estimated confidence scores on input data for each trajectory predictor. The confidence estimator estimates scores for the variational predictor and the odometry predictor with an average error of 1.07 meters and 2.14 meters, respectively. As we wish to leverage the confidence estimator to choose the better predictor, we compute the percentage that it finds the correct better trajectory predictor to be 75\%.

Since the driving domain is very safety-critical, we measure the frequency at which our confidence estimator underestimates the uncertain cases. We define the uncertain case as where both predictors produce errors above 2.54 meters, which is the mean error of a set of complicated cases gathered manually from the data. The percentage of underestimated samples out of all uncertain samples is 18\%, which means our confidence estimator knows 82\% of the time when it is outputting uncertain results. This is crucial in safety-critical systems, especially in shared-control and parallel autonomy systems, as we can change the system's actions according to the certainty of its own perception subsystem.

In comparison, in the cases where the prediction error is less than 2.54 meters, our confidence measure undershoots the error in 94\% of the cases, which means we can trust the confidence estimator 89.2\% of the time among all testing samples with 40\% of uncertain cases and 60\% of certain cases.

\vspace{-1mm}
\subsection{Mixture Predictor}
\label{subsec:mixture_p}
The mixture-of-experts predictor arbitrates between the variational predictor and the odometry predictor, by generating their estimated confidence scores, and picking the prediction that has a higher score. The mixture-based errors are distributed in green as shown in Figure~\ref{fig:temporal_trends}a with an average of 2.33 meters, which is improved by 23\% compared to the average error of the variational predictor and 25\% compared to the average error of the odometry predictor. The \emph{regret} is 0.16 meters, as defined in comparison to an \emph{oracle} arbitrator that knows which prediction is better at every sample. 

\subsection{Temporal Behaviors}

An important aspect of predictors is their behaviors as a function of the prediction horizon. 
We are interested in seeing the performance of the mixture predictor, and gather the mean of L2 prediction errors from different predictors over a predicting horizon from 0.1 seconds to 3 seconds (see Figure~\ref{fig:temporal_trends}b). The gap between the mixture predictor curve and candidate predictor curves indicates that it successfully chooses the good part from them. This can be validated by the temporally-dependent regret as illustrated by the gap between the mixture predictor curve and the oracle predictor curve. Similar to \ref{subsec:conf_est}, we compute the percentage of uncertain cases over different time horizons (see Figure~\ref{fig:temporal_trends}c). Starting at 1.4 seconds, the number of uncertain cases starts to increase quickly, due to the uncertainty increase in driver actions. The percentage of underestimated uncertain cases over total cases never exceeds 10\%, which means that our mixture predictor is aware most of the time when generating predictions with large uncertainty. The underestimated curve starts to fall after 2.5 seconds, as our confidence estimator tends to output larger error estimates over longer horizons and thus is less likely to underestimate the uncertain cases.

% \subsection{Example Visualization}
% \label{subsubsec:visualize}
% In addition to overall statistical results, we want to understand when our model performs well and when it performs badly by visualizing examples with low scores or large errors. Figure \ref{fig:easy} shows the top 10 examples where our model performs well, and most examples are a vehicle driving smoothly on a road with slight curvatures. On the other hand, the top 10 hard cases are where the vehicle makes turns at intersections, as shown in Figure \ref{fig:hard}. Even in the hard cases, we can see our trajectory predictions mostly capture the groundtruth results.

% \subsection{KITTI Evaluation}
% \label{subsubsec:kitti}
% In order to show that our model is not limited to our own dataset, we selected a few cities and residential trips from an open source KITTI dataset \cite{Geiger2013IJRR} to validate our model. Since the KITTI dataset does not have CANBUS data available, we dropped the CANBUS network and retrained our model. Before feeding in the KITTI data, we cropped the image into three parts - middle, left, and right as our model takes images from three front views. 

% [TODO] The prediction results on KITTI is similar to results on our own dataset. Thus our model is general enough to work on any urban driving scenario.

\section{Conclusions}
\label{sec:conclusions}

This paper introduces an uncertainty-aware prediction method for driver trajectory prediction by augmenting a variational deep neural network trajectory predictor with a physics-based predictor using confidence estimates. The variational predictor generates probabilistic predictions from a number of aggregated children networks that process different vehicle sensor data, and the physics-based predictor acts as an additional expert predictor that provides more accurate predictions in simple cases where vehicle accelerations are small. We use a second network to estimate the uncertainty of predictions by different candidate predictors as a function of the predicting horizon. The uncertainty estimates not only help the mixture predictor achieve better accuracy over a horizon up to 3 seconds, but also provide insights on when to use our predictor in the context of shared-autonomy. Given the safety critical nature of autonomous driving tasks, we believe our uncertainty-aware method provides a new direction for making a robust and trustworthy prediction system. In the future, we would like to improve the performance of the variational predictor by exploiting more features from the image data.
%, and further pushes the system to succeed in knowing when it is uncertain about its prediction result.

\bibliographystyle{IEEEtran}
% \IEEEtriggeratref{16}
\bibliography{references}

\end{document}